# TriNER: A Series of Named Entity Recognition Models For Hindi, Bengali & Marathi

**Mohammed Amaan Dhamaskar, Rasika Ransing**

Vidyalankar Institute of Technology
{mohammedamaan.dhamaskar, rasika.ransing}@vit.edu.in

## Abstract

India's rich cultural and linguistic diversity poses various challenges in the domain of Natural Language Processing (NLP), particularly in Named Entity Recognition (NER). NER is a NLP task that aims to identify and classify tokens into different entity groups like Person, Location, Organization, Number, etc. This makes NER very useful for downstream tasks like context-aware anonymization. This paper details our work to build a multilingual NER model for the three most spoken languages in India - Hindi, Bengali & Marathi. We train a custom transformer model and fine tune a few pretrained models, achieving an F1 score of 92.11 for a total of 6 entity groups. Through this paper, we aim to introduce a single model to perform NER and significantly reduce the inconsistencies in entity groups and tag names, across the three languages.

## 1 Introduction

The term "named entity" was first introduced during the sixth Message Understanding Conference [1], where the objective was to identify names of individuals, locations, and organizations in text, later extending to include temporal and numerical expressions. Since then, named entity recognition (NER) has advanced to encompass fine-grained NER, which identifies subcategories within entities [2], and nested NER, which detects entities embedded within others [3]. In the last few years, there has been significant progress in this domain, including varied approaches to perform Named Entity Recognition ranging from the use of Conditional Random Fields (CRF) and Long Short-Term Memory (LSTM) to Embeddings from Language Models (ELMo), and Transformer models. However, Transformer models consistently outperform other approaches for NER, as outlined in [4] and [5], and due to this reason, we focus on using them for our efforts.

The work in NER is rapidly growing, but only a small subset of this progress encompasses Indian regional languages. While the past few years have produced various gold standard datasets and associated models, they suffer from inconsistencies in the targeted entity groups. Additionally, the presence of separate models for each language leads to inefficiencies in use and increased inference costs. To counter this, there are a few multilingual NER models as well, but most of them only recognize Person, Location & Organization.

This paper outlines our effort to overcome these challenges and inconsistencies to develop a multilingual NER model for Hindi, Bengali & Marathi. Our choice of language is driven by the presence of high-quality datasets and the fact that these languages are the 3 most spoken languages of India. We also define a common set of 6 entity classes to ensure consistency across languages. We discuss the steps involved in data curation, implementation of the models and evaluation technique and results used.

## 2 Related Work

In the context of Named Entity Recognition, a substantial body of literature addresses the task across various languages and domains. This section, however, focuses specifically on reviewing existing studies related to Hindi, Marathi, and Bengali.

[5] outlines the IJCNLP NER corpus, which was one of the first efforts in the domain of NER for

South & Southeast Asian languages. It presented a large corpus for various low resource languages which included Hindi & Bengali. [6] utilized this corpus to develop a NER model using CRFs for five Indian languages, including Hindi & Bengali. They achieved substantial precision scores, but the low recall values led to a drop in F1 scores.

FIRE 2014 NER dataset [7] is a NER dataset which includes Hindi among other languages. Using this dataset, [8] demonstrates that training on a combined corpus of labeled data from multiple languages can enhance Named Entity Recognition (NER) performance for Indian languages. [8] also released the IIT Bombay Marathi NER Corpus, to complement other languages in the FIRE 2014 dataset.

[9] released the WikiAnn NER dataset, a silver standard dataset of 10,000 sentences for 282 languages, including Hindi, Marathi & Bengali. [10] utilized transfer learning for 41 languages including Hindi & Bengali, and present their results for Zero Shot and Few Shot transfers.

[11] introduced a NER corpus for Bengali language and utilized it to develop a Bengali NER model using various approaches like Hidden Markov Model (HMM), CRFs & Support Vector Machines (SVM), achieving an F1 score of 91.8. Unfortunately, this dataset was never released in the public domain. [12] & [13] introduced large NER datasets for Bengali language, however [14] highlight inconsistencies in annotations in these datasets. [14] also introduced an extensive Bengali NER corpus, titled B-NER, which included 22,000 sentences, targeting 8 entity groups. Currently, this can be considered as a near-gold standard dataset for Bengali, available publicly. They further benchmarked their dataset by utilizing it to train Bi-LSTMs and fine tune BERT models, achieving an F1 score of 86. They also introduced a manually annotated gold standard dataset comprising 3 entity groups (Person, Location & Organization) for evaluation.

For Hindi, the first gold standard dataset for was the HiNER [15] dataset and corresponding NER models. It comprises over 100,000 manually annotated sentences and 11 entity categories, as opposed to just 3 entity categories in most of the previous works. They achieved a weighted F1 score of 88.78 with all entity classes and 92.22 with 3 - Person, Location & Organization. Similarly for Marathi, the first gold standard dataset was the MahaNER [16] dataset and associated BERT models, which comprises 25,000 manually annotated sentences and 7 entity categories. They achieved F1 scores of 85.3 and 86.8 on the IOB and non-IOB notation versions.

A major breakthrough in the Indian NER landscape was the Naamapadam [17] dataset, which is the largest NER corpora for Indian languages, containing 5.7 million sentences, 3 entity classes and spanning 11 languages which include Hindi, Bengali & Marathi. They also introduced the IndicNER model which achieved F1 scores in the range of 81 to 83 for the three languages. The dataset was generated from the Samanantar parallel corpus [18] by projecting automatically annotated entities from English sentences to their corresponding translations in Indian languages. They also employed certain word alignment methods to ensure consistent tagging across languages. However, a lot of word alignment mistakes can still prevail, which can lead to erroneous tagging of entities.

## 3 Implementation

### 3.1 Dataset Curation

Our aim was to build on the efforts of Hiner, MahaNER & B-NER, and develop a common model to perform NER for the 3 languages. We therefore combined the 3 datasets to produce a master training and validation set with 6 entities: PERSON, LOCATION, ORGANIZATION, NUMEX, TIMEX and MISC. Existing tags of HiNER, MahaNER & B-NER were mapped to our entities as per Table 1. In view of the absence of a NUMEX or equivalent tag in the B-NER dataset, we automatically annotated the dataset using a python script which annotates each token as NUMEX if it can be converted to a float value, unless already annotated as a separate named entity. This allowed us to preserve the annotations of Pin-codes, Date, Time, etc in numerical format. All remaining tags in the source datasets were dropped.

For our training set, we combined the training sets of HiNER & MahaNER and 80% of B-NER training set. This is because the B-NER test set contains only 3 tags (PER, LOC, ORG). The remaining 20% of B-NER training set was combined with 30% of B-NER test set, and the testing and validation sets of HiNER & MahaNER datasets to create our validation set.

| Tags | HiNER | MahaNER | B-NER |
|---|---|---|---|
| **PERSON** | PERSON | NEP | PER |
| **LOCATION** | LOCATION | NEL | GEO |
| **ORGANIZATION** | ORGANIZATION | NEO | ORG |
| **NUMEX** | NUMEX | NEM | Automatically Annotated |
| **TIMEX** | TIMEX | NED, NETI | TIM |
| **MISC** | MISC, LITERATURE, RELIGION | ED | GPE, ART |

Table 1: Mapping of source NER dataset tags to our training and validation set tags

### 3.2 Training Custom Model

We first experimented with building a NER model by training a custom transformer using Keras. We also defined our own vocabulary to tokenize input sequences. The vocabulary comprises 95% of the most common words of each language training set. Combining language wise vocabulary into the master vocabulary allows us to maintain a fair contribution of each language in the master set. Since the least common 5% words in each language usually consist of named entities like names, organization names, etc; excluding them allows us to include out of vocab terms in the training set, thereby ensuring robust performance while handling out of vocab terms during inference.

### 3.3 Fine Tuning Pretrained Models

We also fine-tuned prominent pretrained models, since they are trained on large amounts of data, and yield great results when fine-tuned for downstream tasks. Our selection of models includes XLM Roberta [20], Multilingual Bert (mBERT) [21] and MuRIL [22]. We have selected these models since they are multilingual and support our target languages - Hindi, Marathi & Bengali. Furthermore, these models have produced great results in the prior works mentioned in the literature review.

While fine tuning these models, we experimented with various hyperparameter values and selected the ones which yield best results. While tuning the learning rate and batch size, we consider the following range for batch size { 8, 16, 32, 64 } and the following range for learning rate { 1e-4, 2e-4, 4e-4, 1e-5, 2e-5, 4e-5, 1e-6, 2e-6, 5e-6 }. The learning rate of 2e-5 produced best results, coupled with a batch size of 16 for XLM Roberta and 32 for MuRIL & mBERT. Each model was fine tuned for 5 epochs and we discovered that further training yielded very small and diminishing increments in the models' performance.

## 4 Results & Discussions

### 4.1 Evaluation Method

We employ the use of seqeval, which is a widely used Python framework to evaluate models performing sequence labelling tasks, including NER, Part-of-Speech [POS] tagging, semantic role labeling, etc. For NER, it supports a variety of annotation schemas including the IOB2 scheme used by us.

### 4.2 Evaluation Results

The overall performance of our models is outlined in Table 2. The fine tuned version of XLM-R outperforms all other models in almost all metrics, with an F1-Score of 92.11, achieving state-of-the-art performance for multilingual NER for Indian regional languages. Furthermore, MuRIL has a slightly better and consistent Recall. These results are a product of reduced number of tags as compared to the respective source datasets and rigorous hyperparameter tuning. We also discover that our results are consistent with the finding of [9] - training a model on a combined multilingual corpus enhances NER performance for most, especially low resource languages.

| Metric | Custom Decoder Only Model | mBERT Fine Tuned | MuRIL Fine Tuned | XLM-R Fine Tuned |
|---|---|---|---|---|
| **Accuracy** | 95.31 | 97.53 | 97.73 | **97.81** |
| **Precision** | 76.94 | 90.53 | 91.31 | **91.68** |
| **Recall** | 81.76 | 91.66 | **92.58** | 92.54 |
| **F1** | 79.27 | 91.09 | 91.94 | **92.11** |

Table 2: Overall performance of our models

The tag wise performance of our models is outlined in Table 3. We observe that XLM-R outperforms other models in recognizing most entity classes, which also reflects in the weighted scores. However, we also discover that MuRIL consistently outperforms other models in handling out of vocabulary values. Due to this reason, we believe that the MuRIL based model would be more suitable for real world use, since the differences in performance of these two models is very small.

| Tag | Custom Decoder Only Model | mBERT Fine Tuned | MuRIL Fine Tuned | XLM-R Fine Tuned |
|---|---|---|---|---|
| **PERSON** | 74.85 | 84.57 | **87.17** | 87.02 |
| **LOCATION** | 90.42 | 95.04 | 95.53 | **95.56** |
| **ORGANIZATION** | 50.00 | 77.66 | 79.52 | **80.60** |
| **NUMEX** | 65.32 | 98.36 | 98.35 | **98.72** |
| **TIMEX** | 60.78 | 84.49 | 84.97 | **85.43** |
| **MISC** | 56.16 | 75.55 | **76.29** | 75.81 |

Table 3: Tag Wise F1 Score of our models

### 4.3 Limitations

Since we have combined different datasets to produce our training and validation sets, there might be certain inconsistencies in tagging, like the some of the language sub entities included in the Miscellaneous class. Additionally, the tokenization methods used by the pretrained models can lead to some inconsistencies during inference. This might require the model outputs to be post-processed before being used for downstream tasks like context-aware anonymization.

## 5 Conclusion

This study presents a comprehensive approach to building a multilingual Named Entity Recognition (NER) model for Hindi, Bengali, and Marathi, the three most widely spoken languages in India. By unifying datasets from HiNER, MahaNER, and B-NER, and mapping their entity classes into a consistent schema, we address key inconsistencies in entity group definitions and tag naming conventions across these languages. The fine-tuned XLM-R model demonstrates state-of-the-art performance, achieving an F1 score of 92.11, validating the hypothesis that a unified multilingual corpus enhances NER performance. Additionally, our findings underscore the potential of pre-trained multilingual models such as MuRIL for handling out-of-vocabulary tokens.

While this work makes significant progress in reducing inefficiencies and inconsistencies in Indian language NER, limitations such as tagging variations and tokenization inconsistencies point to areas for further refinement. Future work could focus on expanding the model to include additional Indian languages, and introducing more entity groups for more precise classifications.

### Acknowledgements

A part of this effort was developed under the AI Sprint hosted by the Machine Learning Developer Programs Team at Google and was supported in the form of Colab Credits for training the models. We thank the team for their continued support.